\documentclass[10pt,twocolumn,letterpaper]{article}

\usepackage[pagenumbers]{cvpr} 

\usepackage[dvipsnames]{xcolor}

\definecolor{cvprblue}{rgb}{0.21,0.49,0.74}
\usepackage[pagebackref,breaklinks,colorlinks,citecolor=cvprblue]{hyperref}

\usepackage{url}
\usepackage{mathtools}
\usepackage{bm}
\usepackage{xfrac}
\usepackage{xspace}
\usepackage[makeroom]{cancel}
\usepackage{makecell}
\usepackage{multirow}
\usepackage{subcaption}
\usepackage{color}
\usepackage{xcolor}
\usepackage{colortbl}
\usepackage{lipsum}
\usepackage{ wasysym }
\usepackage{textcomp}
\usepackage[abs]{overpic}
\usepackage{float}

\usepackage{booktabs}
\definecolor{myyellow}{rgb}{1,1, 0.6}
\definecolor{myorange}{rgb}{1, 0.8, 0.6}
\definecolor{myred}{rgb}{1, 0.6, 0.6}
\definecolor{blue-violet}{rgb}{0.54, 0.17, 0.89}
\definecolor{blue-violet}{rgb}{0.54, 0.17, 0.89}
\usepackage{amssymb}

\def\paperID{***} 
\def\confName{CVPR}
\def\confYear{2024}

\title{DynVideo-E: Harnessing Dynamic NeRF for Large-Scale Motion- and View-Change Human-Centric Video Editing}

\author{%
 Jia-Wei Liu{$^{1*}$}, Yan-Pei Cao{$^{3\dag}$}, Jay Zhangjie Wu{$^{1}$}, Weijia Mao{$^{1}$}, Yuchao Gu{$^{1}$}, Rui Zhao{$^{1}$}, \\ Jussi Keppo{$^{2}$}, Ying Shan{$^{3}$}, Mike Zheng Shou{$^{1\dag}$}
 \\\\
 $\textsuperscript{\rm 1}$Show Lab, $\textsuperscript{\rm 2}$National University of Singapore\quad
 $\textsuperscript{\rm 3}$ARC Lab, Tencent PCG
 \\ 
   \vspace{-2em}
}

\begin{document}

\newcommand*{\method}{DynVideo-E}

\twocolumn[{
\maketitle
\vspace{-2em}
\renewcommand\twocolumn[1][]{#1}
\begin{center}
    \centering
    \includegraphics[width=1\linewidth]{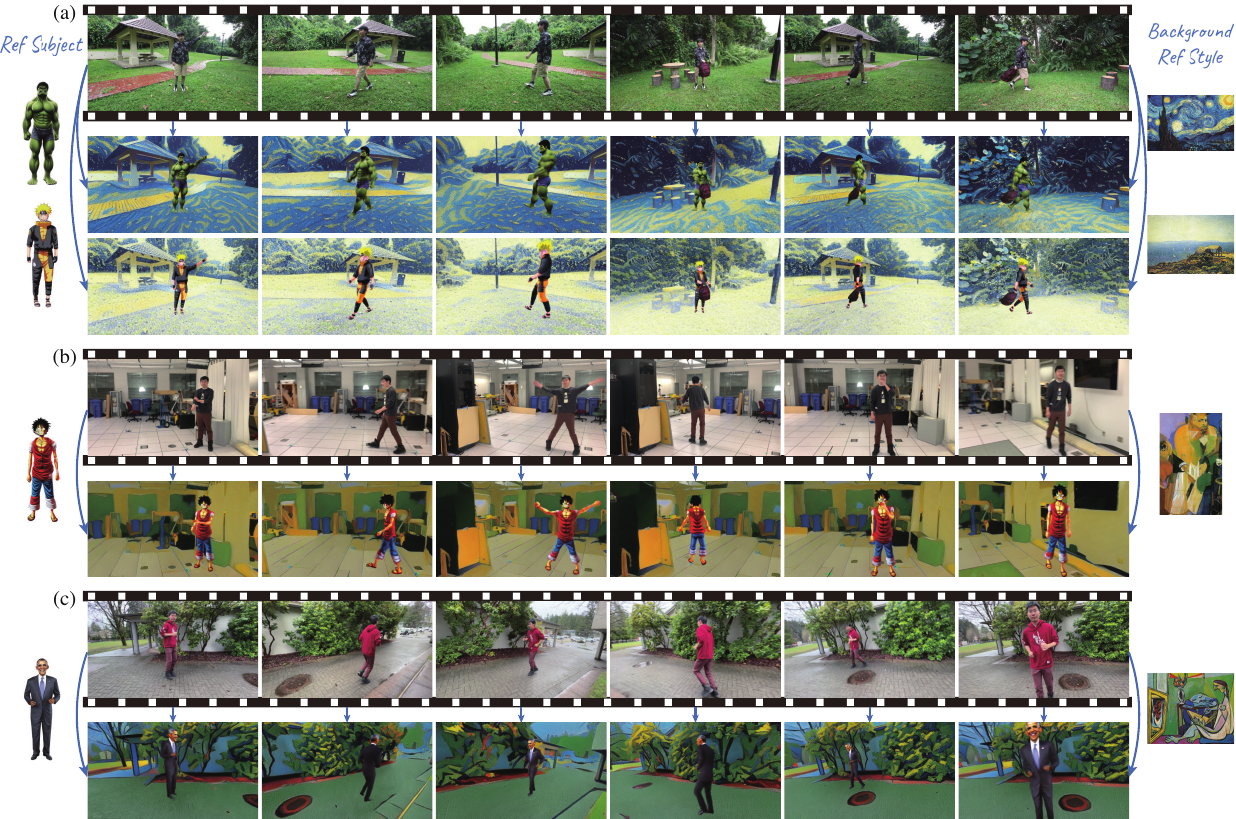}
    \captionof{figure}{Given a reference subject image and a background style image, our \method{} enables highly consistent editing of large-scale motion- and view-change human-centric videos (a-c).}
    \label{fig:teaser}
\end{center}
}]

\renewcommand{\thefootnote}{\fnsymbol{footnote} }{}
\footnotetext[1]{Work is partially done during internship at ARC Lab, Tencent PCG. \\ \indent\,$^{\dag}$ Corresponding Authors.}

\begin{abstract}

Despite recent progress in diffusion-based video editing, existing methods are limited to short-length videos due to the contradiction between long-range consistency and frame-wise editing. Prior attempts to address this challenge by introducing video-2D representations encounter significant difficulties with large-scale motion- and view-change videos, especially in human-centric scenarios. To overcome this, we propose to introduce the dynamic Neural Radiance Fields (NeRF) as the innovative video representation, where the editing can be performed in the 3D spaces and propagated to the entire video via the deformation field. To provide consistent and controllable editing, we propose the image-based video-NeRF editing pipeline with a set of innovative designs, including multi-view multi-pose Score Distillation Sampling (SDS) from both the 2D personalized diffusion prior and 3D diffusion prior, reconstruction losses, text-guided local parts super-resolution, and style transfer. Extensive experiments demonstrate that our method, dubbed as~\method{}, significantly outperforms SOTA approaches on two challenging datasets by a large margin of $50\%\sim95\%$ for human preference. Code will be released at~\url{https://showlab.github.io/DynVideo-E/}.

\end{abstract}

\section{Introduction}

The remarkable success of powerful image diffusion models~\cite{rombach2022high} has sparked considerable interests in extending them to support video editing~\cite{wu2022tune}. Despite promising, it presents significant challenges in maintaining high temporal consistency. To tackle this problem, existing diffusion-based video editing approaches have evolved to extract and incorporate various correspondences from source videos into the frame-wise editing process, including attention maps~\cite{qi2023fatezero, liu2023video}, spatial maps~\cite{yang2023rerender, zhao2023controlvideo}, optical flows and nn-fields~\cite{geyer2023tokenflow}. While these works have demonstrated enhanced temporal consistency of editing results, the inherent contradiction between long-range consistency and frame-wise editing limits these methods to short-length videos with small motions and viewpoint changes. 

Another line of research seeks to introduce intermediate video-2D representations to degrade video editing to image editing, such as decomposing videos using the layered neural atlas~\cite{kasten2021layered} and mapping spatial-temporal contents to 2D UV maps. As such, editing can be performed on a single frame~\cite{huang2023inve, lee2023shape} or on the atlas itself~\cite{chai2023stablevideo, couairon2023videdit, bar2022text2live}, with the edited results consistently propagating to other frames. More recently, CoDeF~\cite{ouyang2023codef} proposes the 2D hash-based canonical image coupled with a 3D deformation field to further improve the video representative capability. However, these approaches are 2D representations of video contents, and thus they encounter significant difficulties in representing and editing videos with large-scale motion and viewpoint changes, especially in human-centric scenarios. 

This motivates us to introduce the video-3D representation for large-scale motion- and view-change human-centric video editing. Recent advances in dynamic NeRF~\cite{liu2023hosnerf,weng2022humannerf,jiang2022neuman} show that the 3D dynamic human space coupled with the human pose guided deformation field can effectively reconstruct single human-centric videos with large motions and viewpoints changes. Therefore, in this paper, we propose~\method{} that for the first time introduces the dynamic NeRF as the innovative video representation for challenging human-centric video editing. Such a video-NeRF representation effectively aggregates the large-scale motion- and view-change video information into a 3D background space and a 3D dynamic human space through the human pose guided deformation field, and thus the editing can be performed in the 3D spaces and propagated to the entire video via the deformation field.

To provide consistent and controllable editing, we propose the image-based video-NeRF editing pipeline with a set of effective designs. These include 1) reconstruction losses on the reference image under reference human pose and camera viewpoint to inject subject contents from the reference image to the 3D dynamic human space. 2) To improve the 3D consistency and animatability of the edited 3D dynamic human space, we design a multi-view multi-pose Score Distillation Sampling (SDS) from both the 2D personalized diffusion prior and 3D diffusion prior, as well as a set of training strategies under various human pose and camera pose configurations. 3) To improve the resolution and geometric details of 3D dynamic human space, we utilize the text-guided local parts zoom-in super-resolution with $7$ semantic body regions augmented with view conditions. 4) We employ a style transfer module to transfer the reference style to our 3D background model. After training, our video-NeRF model can render highly consistent videos along source video viewpoints by propagating the edited contents through the deformation field, and it can achieve 360{\textdegree} free-viewpoint high-fidelity novel view synthesis for edited dynamic scenes.

We extensively evaluate our~\method{} on HOSNeRF~\cite{liu2023hosnerf} and NeuMan~\cite{jiang2022neuman} dataset with $24$ editing prompts on $11$ challenging dynamic human-centric videos. As shown in Fig.~\ref{fig:teaser}, our \method{} generates photorealistic video editing results with very high temporal consistency, and significantly outperforms SOTA approaches by a large margin of $50\%\sim95\%$ in terms of human preference.

To summarize, the major contributions of our paper are:

\begin{itemize}
\item
We present a novel framework of \method{} that for the first time introduces the dynamic NeRF as the innovative video representation for large-scale motion- and view-change human-centric video editing.
\item
We propose a set of effective designs and training strategies for the image-based 3D dynamic human and static background space editing in our video-NeRF model.
\item
\method{} significantly outperforms SOTA approaches on two challenging datasets by a large margin of $50\%\sim95\%$ for human preference and achieves high-fidelity free-viewpoint novel view synthesis for edited scenes.

\end{itemize}

\section{Related Work}

\subsection{Diffusion-based Video Editing}
Thanks to the power of diffusion models, prior works have extended their support to video editing~\cite{wu2022tune,qi2023fatezero} and generation~\cite{blattmann2023align,zhang2023show1}. Pioneer Tune-A-Video~\cite{wu2022tune} inflates the image diffusion with cross-frame attention and fine-tunes the source video, aiming to implicitly learn the source motion and transfer it to the target video. Although Tune-A-Video~\cite{wu2022tune} demonstrates versatility across different video editing applications, it exhibits inferior temporal consistency. Subsequent works extract various correspondences from the source video and employ them to improve temporal consistency. FateZero~\cite{qi2023fatezero} and Video-P2P~\cite{liu2023video} extract the cross- and self-attention from the source video to control the spatial layout. Rerender-A-Video~\cite{yang2023rerender}, ControlVideo~\cite{zhao2023controlvideo}, and TokenFlow~\cite{geyer2023tokenflow} extract and align optical flows, spatial maps, and nn-fields from the source video, resulting in improved consistency of editing results. Although these works have shown promising results, they are typically used in short-form video editing scenarios with small-scale motions and view changes. 

Another line of video editing work relies on a powerful video representation, namely, the layered neural atlas~\cite{kasten2021layered}, as an intermediate editing representation. The layered neural atlas factorizes the input video using a layered presentation and maps the subject and background of all frames to 2D UV maps. Once the layered neural atlas is learned, editing can occur either on keyframes~\cite{huang2023inve, lee2023shape} or on the atlas itself~\cite{chai2023stablevideo, couairon2023videdit, bar2022text2live}, and the editing results consistently propagate to other frames. CoDeF~\cite{ouyang2023codef} incorporates the 3D deformation field with the 2D hash-based canonical image to further improve the video representative capability. However, both the layered neural atlas~\cite{kasten2021layered} and canonical image~\cite{ouyang2023codef} are pseudo-3D representations of video contents, and they encounter difficulties in reconstructing videos with large-scale motion and viewpoint changes.

\subsection{Dynamic NeRFs}

Remarkable progress has been made in the field of novel view synthesis since the introduction of Neural Radiance Fields (NeRF)~\cite{mildenhall2021nerf}. Subsequent studies have extended it to reconstruct dynamic NeRFs from monocular videos by either learning a deformation field that maps sampled points from the deformed space to the canonical space~\cite{pumarola2021d,park2021nerfies,park2021hypernerf,tretschk2021non} or building 4D spatio-temporal radiance fields~\cite{xian2021space,li2021neural,gao2021dynamic}. Other studies have introduced voxel grids~\cite{liu2022devrf,fang2022fast,song2022nerfplayer} or planar representations~\cite{fridovich2023k,cao2023hexplane} to improve the training efficiency of dynamic NeRFs. While these approaches have shown promising results, they are limited to short videos with simple deformations. Another series of work focus on human modelling and leverage estimated human pose priors~\cite{peng2021neural,weng2022humannerf} to reconstruct dynamic humans with complex motions. Recently, NeuMan~\cite{jiang2022neuman} reconstructs the dynamic human NeRF together with static scene NeRF to model human-centric scenes. HOSNeRF~\cite{liu2023hosnerf} further proposes to represent the complex human-object-scene with the state-conditional dynamic human model and unbounded background model, achieving 360{\textdegree} free-viewpoint renderings from single videos. In contrast, we aim to introduce the dynamic NeRF as the innovative video-NeRF representation for human-centric video editing.

\subsection{NeRF-based Editing and Generation}

Since the introduction of diffusion models, text-guided 3D NeRF editing and generation has evolved from CLIP-based~\cite{jain2022zero,wang2023nerf,hong2022avatarclip} to 2D diffusion-based~\cite{zhuang2023dreameditor,mikaeili2023sked,sella2023vox,li2023focaldreamer,kolotouros2023dreamhuman} methods. SINE~\cite{bao2023sine} supports editing a local region of static NeRF from a single view by delivering edited contents to multi-views through pretrained NeRF priors. ST-NeRF~\cite{zhang2021editable} presents a spatiotemporal neural layered radiance representation to represent dynamic scenes with layered NeRFs, and it can achieve simple editing such as affine transform or duplication by manipulating the NeRF layer. However, it requires $16$ cameras to capture a dynamic scene and cannot edit the contents of layered NeRFs. Subsequent works such as Control4D~\cite{shao2023control4d} and Dyn-E~\cite{zhang2023dyn} propose to edit the contents of dynamic NeRFs. However, Control4D~\cite{shao2023control4d} is limited to human-only scenes with small motions and short video length, while Dyn-E~\cite{zhang2023dyn} only supports editing the local appearance with explicit user manipulation. 

\section{Method}

\begin{figure*}
\begin{centering}
\includegraphics[width=1\linewidth]{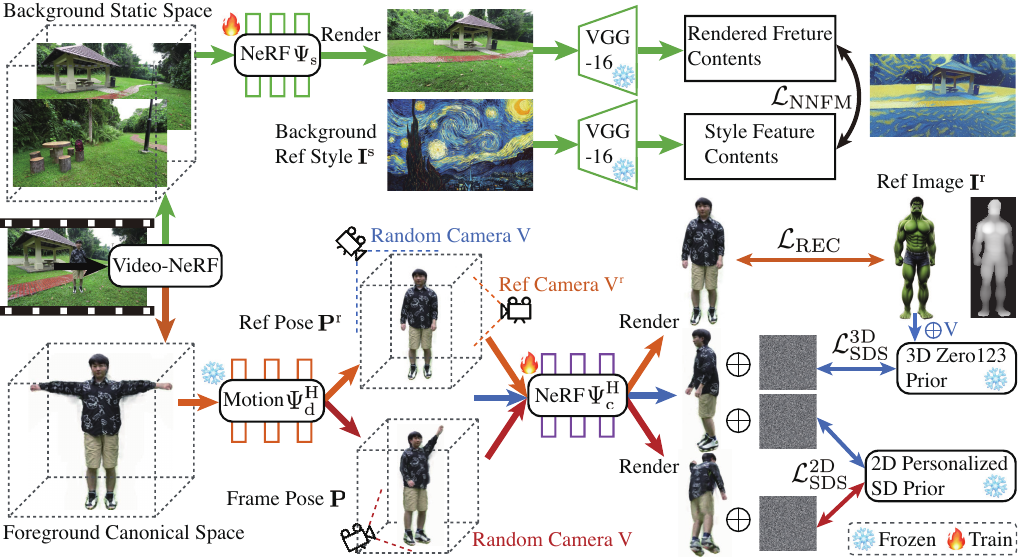}
\par\end{centering}
\caption{\label{fig:pipeline}\textbf{Overview of \method{}.} (1) Our \textbf{video-NeRF model} represents the input video as a 3D dynamic human space coupled with the deformation field and a 3D static background space. (2) \textbf{Orange flowchart:} Given the reference subject image, we edit the animatable 3D dynamic human space under multi-view multi-pose configurations by leveraging reconstruction losses, 2D personalized diffusion priors, 3D diffusion priors, and local parts super-resolution. (3) \textbf{Green flowchart:} A style transfer loss in feature spaces is utilized to transfer the reference style to our 3D background model. (4) \textbf{Edited videos} can be accordingly rendered by volume rendering in the edited video-NeRF model under source video camera poses, and we can also achieve high-fidelity free-viewpoint renderings of edited dynamic scenes.}
\vspace{-5mm}
\end{figure*}

\subsection{Video-NeRF Model}

\noindent\textbf{Motivation.} Given single videos with large viewpoint changes, intricate scene contents, and complex human motions, we seek to represent such videos using dynamic NeRFs for video editing. HOSNeRF~\cite{liu2023hosnerf} has been recently proposed to reconstruct the dynamic neural radiance fields for dynamic human-object-scene interactions from a single monocular in-the-wild video and achieves new SOTA performances. It proposes the state-conditional 3D dynamic human-object model and 3D background model to separately represent the dynamic human-object and static background. Therefore, we harness HOSNeRF~\cite{liu2023hosnerf} as our video-NeRF model to represent large-scale motion- and view-change human-centric videos that consists of dynamic humans, dynamic objects, and static backgrounds. Since our goal is to edit the dynamic human and unbounded background while keep the interacted objects unchanged, we utilize the original reconstructed HOSNeRF model to keep interacted objects, and simplify HOSNeRF~\cite{liu2023hosnerf} to HSNeRF by removing the object state designs for video editing. Therefore, our video-NeRF model consists of a dynamic human model $\Psi^{\mathrm{H}}$ and a static scene model $\Psi^{\mathrm{S}}$.

\noindent\textbf{3D Dynamic Human Model $\Psi^{\mathrm{H}}$} aggregates the dynamic information across all video frames into a 3D canonical human space $\Psi_{\mathrm{c}}^{\mathrm{H}}$ that maps 3D points to color $\mathbf{c}$ and density $d$, and a human pose-guided deformation field $\Psi_{\mathrm{d}}^{\mathrm{H}}$ that maps deformed points $\mathbf{x}^{i}_{\mathrm{d}}$ from the deformed space at frame $i$ to canonical points $\mathbf{x}^{i}_{\mathrm{c}}$ in the canonical space ($i$ omitted for simplicity).
\begin{equation}
\Psi_{\mathrm{c}}^{\mathrm{H}}\left(\gamma\left(\mathbf{x}_{\mathrm{c}}\right)\right)\longmapsto\left(\mathbf{c},\,d\right),\;\Psi_{\mathrm{d}}^{\mathrm{H}}\left(\mathbf{x}_{\mathrm{d}},\,\mathcal{J},\,\mathcal{R}\right)\longmapsto\left(\mathbf{x}_{\mathrm{c}}\right),\,\label{eq:canonical}
\end{equation}
where $\gamma\left(\mathbf{x}\right)$ is the standard positional encoding function, and $\mathcal{J}=\left\{ \mathbf{J}_{i}\right\} \,$ and $\mathcal{R}=\left\{ \boldsymbol{\omega}_{i}\right\} \,$ are 3D human joints and local joint axis-angle rotations, respectively.

Following HOSNeRF~\cite{liu2023hosnerf} and HumanNeRF~\cite{weng2022humannerf}, we decompose the deformation field $\Psi_{\mathrm{d}}^{\mathrm{H}}$ into a coarse human skeleton-driven deformation $\Psi_{\mathrm{d}}^{\mathrm{H, coarse}}$ and a fine non-rigid deformation conditioned on human poses $\Psi_{\mathrm{d}}^{\mathrm{H, fine}}$:
\begin{equation}
\mathbf{x}_{\mathrm{c}}^{\prime}=\Psi_{\mathrm{d}}^{\mathrm{H, coarse}}\left(\mathbf{x}_{\mathrm{d}},\,\mathcal{J},\,\mathcal{R}\right),\;\mathbf{x}_{\mathrm{c}}=\mathbf{x}_{\mathrm{c}}^{\prime}+\Psi_{\mathrm{d}}^{\mathrm{H, fine}}\left(\mathbf{x}_{\mathrm{c}}^{\prime},\,\mathcal{R}\right)\,.
\end{equation}
\noindent\textbf{3D Static Scene Model $\Psi^{\mathrm{S}}$} aggregates intricate static scene contents into a Mip-NeRF 360~\cite{barron2022mip} space that maps contracted Gaussian parameters $(\boldsymbol{\hat{\mu}}, \boldsymbol{\hat{\Sigma}})$ to color $\mathbf{c}$ and density $d$.
\begin{equation}
\Psi_{\mathrm{s}}\left(\hat{\gamma}\left(\boldsymbol{\hat{\mu}},\,\boldsymbol{\hat{\Sigma}}\right)\right)\longmapsto\left(\mathbf{c},\,\sigma\right)\,,
\end{equation}
where $\hat{\gamma}$ is the integrated positional encoding (IPE)~\cite{barron2022mip}:
\begin{equation}
\resizebox{2.95in}{!}{$
\hat{\gamma}(\boldsymbol{\hat{\mu}}, \boldsymbol{\hat{\Sigma}}) 
= \Bigg\{ \begin{bmatrix} \sin(2^\ell \boldsymbol{\hat{\mu}}) \exp\left(-2^{2\ell - 1} \operatorname{diag}\left(\boldsymbol{\hat{\Sigma}}\right)\right)
\\ \cos(2^\ell \boldsymbol{\hat{\mu}}) \exp\left(-2^{2\ell - 1} \operatorname{diag}\left(\boldsymbol{\hat{\Sigma}}\right)\right)\end{bmatrix} \Bigg\}_{\ell=0}^{L-1}\,.
$}\label{eq:IPE}
\end{equation}
To obtain the contracted Gaussian parameters, we first split the casted rays into a set of intervals $T_i = [t_i, t_{i+1})$ and compute their corresponding conical frustums' mean and covariance as $(\boldsymbol{\mu}, \boldsymbol{\Sigma}) = \mathbf{r}(T_i)$~\cite{barron2022mip}. Then we adopt the contraction function $f\left(\mathbf{x}\right)$ proposed in Mip-NeRF 360~\cite{barron2022mip} to distribute distant points proportionally to disparity, and parameterize the Gaussian parameters for unbounded scenes as follows,
\begin{equation}
f\left(\mathbf{x}\right)=\left\{ \begin{array}{cc}
\mathbf{x} & \left\Vert \mathbf{x}\right\Vert \leq1\\
\left(2-\frac{1}{\left\Vert \mathbf{x}\right\Vert }\right)\left(\frac{\mathbf{x}}{\left\Vert \mathbf{x}\right\Vert }\right) & \left\Vert \mathbf{x}\right\Vert >1
\end{array}\right.,\,
\end{equation}
and $f\left(\mathbf{x}\right)$ is applied to $(\boldsymbol{\mu}, \boldsymbol{\Sigma})$ to obtain contracted Gaussian parameters:
\begin{equation}
\left(\boldsymbol{\hat{\mu}},\,\boldsymbol{\hat{\Sigma}}\right)=\left(f\left(\boldsymbol{\mu}\right),\,\mathbf{J}_{f}\left(\boldsymbol{\mu}\right)\boldsymbol{\Sigma}\mathbf{J}_{f}\left(\boldsymbol{\mu}\right)^{\mathrm{T}}\right)\,,
\end{equation}
\noindent where $\mathbf{J}_{f}(\boldsymbol{\mu})$ is the Jacobian of $f$ at $\boldsymbol{\mu}$.

\noindent{\bf Video-NeRF Optimization.}
Given single videos with camera poses calibrated using COLMAP~\cite{schoenberger2016sfm,schoenberger2016mvs}, our video-NeRF model is trained by minimizing the difference between the rendered pixel colors and ground-truth pixel colors. To render pixel colors, we shoot rays and query the scene properties in the 3D dynamic human model and scene model, and re-order all sampled properties based on their distances from the camera center. Then, the pixel color can be calculated through the volume rendering~\cite{mildenhall2021nerf}:
\begin{eqnarray}\label{eq:render}
\hat{\mathbf{C}}\left(\mathbf{r}\right) =  \sum_{i=1}^{N}T_{i}\left(1-e^{-\sigma_{i}\delta_{i}}\right)\mathbf{c}_{i},\quad T_{i}  = e^{-\sum_{j=1}^{i-1}\sigma_{j}\delta_{j}}\,.
\end{eqnarray}
Following HOSNeRF~\cite{liu2023hosnerf}, we optimize our video-NeRF representation by minimizing the photometric MSE loss, patched-based perceptual LPIPS~\cite{zhang2018unreasonable} loss, and the regularization losses proposed by Mip-NeRF 360~\cite{barron2022mip} to avoid background collapse, deformation cycle consistency, and indirect optical flow supervisions. Please refer to HOSNeRF~\cite{liu2023hosnerf} for more details.

\subsection{Image-based Video-NeRF Editing}

\noindent\textbf{Motivation.} Previous video editing works~\cite{khachatryan2023text2video,ouyang2023codef,yang2023rerender,bar2022text2live,chai2023stablevideo} primarily describe intended editing through text prompts. However, finer-grained details and the concept's identity are better conveyed through reference images. To this end, we focus on the image based editing for finer and direct controllability. As shown in Fig.~\ref{fig:pipeline}, our video-NeRF model represents the large-scale motion- and view-change human-centric video with a 3D dynamic human space and a 3D background space. Therefore, to better disentangle the foreground and background editing, we propose to edit the 3D dynamic human space with both the reference subject image and its text description, and edit the background static space with the reference style image.

\subsubsection{Image-based 3D Dynamic Human Editing}

\noindent\textbf{Challenges.} Consistent and high-quality image-based video editing requires the edited 3D dynamic human space to 1) keep the subject contents of the reference image; 2) animatable by the human poses from the source video; 3) consistent along large-scale motion and viewpoint changes; and 4) high-resolution with fine details. To address these challenges, we design a set of strategies below.

\noindent\textbf{Reference Image Reconstruction Loss.} We utilize a reference subject image $\mathbf{I}^\mathrm{r}$ to provide finer identity controls and allow for personalized human editing. To ensure that the reference image has a similar human pose with respect to the source human, we leverage ControlNet~\cite{zhang2023adding} to generate the reference subject image conditioned on a source human pose $\mathbf{P}^\mathrm{r}$, as exampled in Fig.~\ref{fig:pipeline}. Then, we use a pretrained monocular depth estimator~\cite{ranftl2020towards} to estimate the pseudo depth $\mathbf{D}^\mathrm{r}$ of reference subject and use SAM~\cite{kirillov2023segment} to obtain its mask $\mathbf{M}^\mathrm{r}$. During training, we assume the reference image viewpoint to be the front view (Ref Camera $\mathbf{V}^\mathrm{r}$ in Fig.~\ref{fig:pipeline}) and render the subject image $\hat{\mathbf{I}}^\mathrm{r}$ driven by the source human pose $\mathbf{P}^\mathrm{r}$ at $\mathbf{V}^\mathrm{r}$ under our video-NeRF representation. We additionally compute the rendered mask $\hat{\mathbf{M}}^\mathrm{r}$ and depth $\hat{\mathbf{D}}^\mathrm{r}$ at $\mathbf{V}^\mathrm{r}$ by integrating the volume density and sampled distances along the ray of each pixel. Following Magic123~\cite{qian2023magic123}, we supervise our framework at $\mathbf{V}^\mathrm{r}$ viewpoint using the mean squared error (MSE) loss on the reference image and mask, as well as the normalized negative Pearson correlation on the pseudo depth map.
\begin{flalign}
\mathcal{L}_{\mathrm{REC}}=\lambda_{\mathrm{rgb}}\left\Vert \mathbf{M}\odot\left(\hat{\mathbf{I}}^{\mathrm{r}}-\mathbf{I}^{\mathrm{r}}\right)\right\Vert _{2}^{2}+\lambda_{\mathrm{mask}}\left\Vert \hat{\mathbf{M}}^{\mathrm{r}}-\mathbf{M}^{\mathrm{r}}\right\Vert _{2}^{2}\nonumber \\
+\frac{1}{2}\lambda_{\mathrm{depth}}\left(1-\frac{\mathrm{cov\left(\mathbf{M}^{r}\odot\mathbf{D}^{r},\,\mathbf{M}^{r}\odot\hat{\mathbf{D}}^{r}\right)}}{\mathrm{\sigma\left(\mathbf{M}^{r}\odot\mathbf{D}^{r}\right)}\mathrm{\sigma\left(\mathbf{M}^{r}\odot\hat{\mathbf{D}}^{r}\right)}}\right)
\end{flalign}
where $\lambda_{\mathrm{rgb}},\,\lambda_{\mathrm{mask}},\,\lambda_{\mathrm{depth}}$ are the loss weights, $\odot$ is the Hadamard product, $\text{cov}(\cdot)$ is the covariance, and $\sigma(\cdot)$ is the standard deviation. 

\noindent\textbf{Score Distillation Sampling (SDS) from 3D Diffusion Prior.} Although $\mathcal{L}_{\mathrm{REC}}$ can provide supervision on the reference image contents, it only works on the source human pose $\mathbf{P}^\mathrm{r}$ at the reference view $\mathbf{V}^\mathrm{r}$. To provide more 3D supervision from the reference image, we utilize the Zero-1-to-3~\cite{Zero-1-to-3} pretrained on the Objaverse-XL~\cite{objaverseXL} as the 3D diffusion prior to distill the inherent 3D geometric and texture information from the reference image using the SDS loss~\cite{poole2022dreamfusion}. Given the 3D diffusion model $\phi$ with the noise prediction network $\epsilon_{\phi}(\cdot)$, the SDS loss works by directly minimizing the injected noise $\epsilon$ added to the encoded rendered images $\mathbf{I}$ and the predicted noise. Therefore, we render images $\mathbf{I}$ from the 3D dynamic human space driven by the source human pose $\mathbf{P}^\mathrm{r}$ at random camera viewpoints $\mathbf{V}=\left[\mathbf{R},\,\mathbf{T}\right]$, and the SDS loss of Zero-1-to-3~\cite{Zero-1-to-3} can be computed with the reference image $\mathbf{I}^\mathrm{r}$ and the camera pose $\left[\mathbf{R},\,\mathbf{T}\right]$ as conditions: 
\begin{flalign}
&\nabla_{\theta}\mathcal{L}_{\mathrm{SDS}}^{\mathrm{3D}}\left(\phi,F_{\theta}\right)=\nonumber \\
&\lambda_{\mathrm{3D}}\cdot\mathbb{E}_{t,\epsilon}\left[w\left(t\right)\left(\epsilon_{\phi}\left(\mathbf{z}_{t};\mathbf{I}^{\mathrm{r}},t,\mathbf{R},\mathbf{T}\right)-\epsilon\right)\frac{\partial\mathbf{I}}{\partial\theta}\right]
\end{flalign}
where $\mathbf{z}_t$ is the noised latent image by injecting a random Gaussian noise of level $t$ to the encoded rendered images $\mathbf{I}$. $w(t)$ is a weighting function that depends on the noise level $t$. $\theta$ is the optimizable parameters of our \method{}.

\noindent\textbf{SDS from 2D Personalized Diffusion Prior.} The reference image guided supervisions above are limited to edit the 3D human space driven only by the source human pose $\mathbf{P}^\mathrm{r}$, and thus are not sufficient to produce a satisfactory 3D dynamic human space that can be animated by the frame human poses from source videos. To this end, we further animate the 3D dynamic human space with the frame human poses $\mathbf{P}$ from the source video and render images $\mathbf{I}$ at random camera poses $\mathbf{V}$, and we further propose to use the 2D text-based diffusion prior~\cite{rombach2022high} to guide these rendered views. However, naively using the 2D diffusion prior hinders the personalization contents learned from the reference image because the 2D diffusion prior tends to imagine the subject contents purely from text descriptions, as valiated in Fig.~\ref{fig:ablation}. To solve this problem, we further propose to use 2D personalized diffusion prior that is first finetuned on the reference image using Dreambooth-LoRA~\cite{ruiz2023dreambooth,hu2021lora}. To generate more inputs for Dreambooth-LoRA, we augment the reference image with random backgrounds and use Magic123~\cite{qian2023magic123} to augment reference image with multiple views. With such Dreambooth-LoRA finetuned 2D personalized diffusion prior $\phi^{\prime}$ with the noise prediction network $\epsilon_{\phi^{\prime}}(\cdot)$, we further employ the 2D SDS loss to supervise the rendered images $\mathbf{I}$ of the 3D dynamic human space animated by the human poses from the source video and rendered at random camera poses.
\begin{equation}
\nabla_{\theta}\mathcal{L}_{\mathrm{SDS}}^{\mathrm{2D}}\left(\phi^{\prime},F_{\theta}\right)=\lambda_{\mathrm{2D}}\mathbb{E}_{t,\epsilon}\left[w\left(t\right)\left(\epsilon_{\phi^{\prime}}\left(\mathbf{z}_{t};y,t\right)-\epsilon\right)\frac{\partial\mathbf{I}}{\partial\theta}\right]
\end{equation}
where $y$ is the text embedding.

\noindent\textbf{Text-guided Local Parts Super-Resolution.} Due to the GPU memory limitation, our \method{} is trained with $\left(128\times128\right)$ resolutions, resulting in coarse geometry and blurry textures. To solve this problem, inspired by DreamHuman~\cite{kolotouros2023dreamhuman}, we utilize the text-guided local parts super-resolution to render and supervise the local parts of zoom-in humans, which improves the effective resolution. Because our dynamic human model is a human pose-driven 3D canonical space under the ``T-pose'' configuration, we can accurately render the zoom-in local human body parts by directly locating the camera close to the corresponding parts. Specifically, we utilize $7$ semantic regions: full body, head, upper body, midsection, lower body, left arm, and right arm, and we accordingly modify the input text prompt with body parts and additionally augment these prompts with view-conditional prompts: front view, side view, and back view. Since it is difficult to track the arm's position under all human poses due to occlusions, we only zoom in on the arms under the ``T-pose''. We provide 8 visualization examples of text-guided local parts super-resolution sampled during training in supplementary materials.

\noindent\textbf{Dynamic Objects.} For human-centric videos with dynamic interacted objects, we utilize the original reconstructed HOSNeRF model to render the interacted objects. During inference, we query the original HOSNeRF model for the rays within object masks, and query the edited video-NeRF model for the rays outside the object masks. As such, we can maintain the dynamic objects in our edited videos.

\subsubsection{Image-based 3D Background Editing}

We aim at transferring the artistic features of an arbitrary 2D reference style image to our 3D unbounded scene model. As shown in the green flowchart of Fig.~\ref{fig:pipeline}, we take inspiration from ARF~\cite{zhang2022arf} and adopt its nearest neighbor feature matching (NNFM) style loss to transfer the semantic visual details from the 2D reference image $\mathbf{I}^{\textrm{s}}$ to our 3D background model $\Psi^{\mathrm{S}}$. We additionally utilize the deferred back-propagation~\cite{zhang2022arf} to directly optimize our model on full-resolution renderings. Specifically, we render the background images $\mathbf{I}$ and extract the VGG~\cite{simonyan2014very} feature maps $\mathbf{F}$ and $\mathbf{F}^{\textrm{s}}$ for $\mathbf{I}$ and $\mathbf{I}^{\textrm{s}}$, respectively, and $\mathcal{L}_{\mathrm{NNFM}}$ minimizes the cosine distance between the rendered feature map and its nearest neighbor in the reference feature map.
\begin{flalign}
&\mathcal{L}_{\mathrm{NNFM}}=\lambda_{\mathrm{NNFM}}\cdot\frac{1}{N}\sum_{i,j}\underset{i^{\prime},j^{\prime}}{\mathrm{min}}D\left(\mathbf{F}\left(i,j\right),\,\mathbf{F}^{\mathrm{s}}\left(i^{\prime},j^{\prime}\right)\right)\,,\\
&D\left(\mathbf{v}_{1},\,\mathbf{v}_{2}\right)=1-\frac{\mathbf{v}_{1}^{\mathrm{T}}\mathbf{v}_{2}}{\sqrt{\mathbf{v}_{1}^{\mathrm{T}}\mathbf{v}_{1}\mathbf{v}_{2}^{\mathrm{T}}\mathbf{v}_{2}}}\,.
\end{flalign}
To prevent the 3D scene model from deviating much from the source contents, we also add an additional L2 loss penalizing the difference between $\mathbf{F}$ and $\mathbf{F}^{\textrm{s}}$~\cite{zhang2022arf}.

\subsection{Training Objectives}

The training of \method{} consists of 2 stages. Firstly, we reconstruct our video-NeRF model on the source video. Secondly, we edit the 3D dynamic human space and 3D unbounded scene space given reference images and text prompts. After training, we render the edited videos using our edited video-NeRF model along source video camera viewpoints, and we also achieve free-viewpoint renderings.

\noindent\textbf{Multi-view Multi-pose Training for 3D Dynamic Human Spaces.} As shown in Fig.~\ref{fig:pipeline}, we design a multi-view multi-pose training process with three conditions during training. 
\begin{itemize}
\item Orange flowchart in Fig.~\ref{fig:pipeline}: only $\mathcal{L}_{\mathrm{REC}}$ is used to supervise the rendered images under source human pose $\mathbf{P}^\mathrm{r}$ at the reference camera view $\mathbf{V}^\mathrm{r}$.
\item Blue flowchart in Fig.~\ref{fig:pipeline}: $\mathcal{L}_{\mathrm{SDS}}^{\mathrm{3D}}$ and $\mathcal{L}_{\mathrm{SDS}}^{\mathrm{2D}}$ are jointly used to supervise the rendered images under source human pose $\mathbf{P}^\mathrm{r}$ at random camera view $\mathbf{V}$.
\item Red flowchart in Fig.~\ref{fig:pipeline}: only $\mathcal{L}_{\mathrm{SDS}}^{\mathrm{2D}}$ is used to supervise the rendered images under the frame human pose $\mathbf{P}$ from the source video at random camera view $\mathbf{V}$.
\end{itemize}
\section{Experiments}

\noindent\textbf{Dataset.} To evaluate our \method{} on both long and short videos, we utilize HOSNeRF~\cite{liu2023hosnerf} dataset with $\left[300,\,400\right]$ frames per video and NeuMan~\cite{liu2023hosnerf} dataset with $\left[30,\,90\right]$ frames per video, all at a resolution of $\left(1280\times720\right)$. In total, we design $24$ editing prompts on $11$ challenging dynamic human-centric videos to evaluate our \method{} and all SOTA Approaches.

\subsection{Comparisons with SOTA Approaches}

\noindent\textbf{Baselines.} We compare our method against five SOTA approaches, including Text2Video-Zero~\cite{khachatryan2023text2video}, Rerender-A-Video~\cite{yang2023rerender}, Text2LIVE~\cite{bar2022text2live}, StableVideo~\cite{chai2023stablevideo}, and CoDeF~\cite{ouyang2023codef}. We utilize Midjourney~\footnote{https://www.midjourney.com/} to generate the text descriptions of the reference images to train these baselines.

\begin{figure*}[!h]
\begin{centering}
\includegraphics[width=1\linewidth]{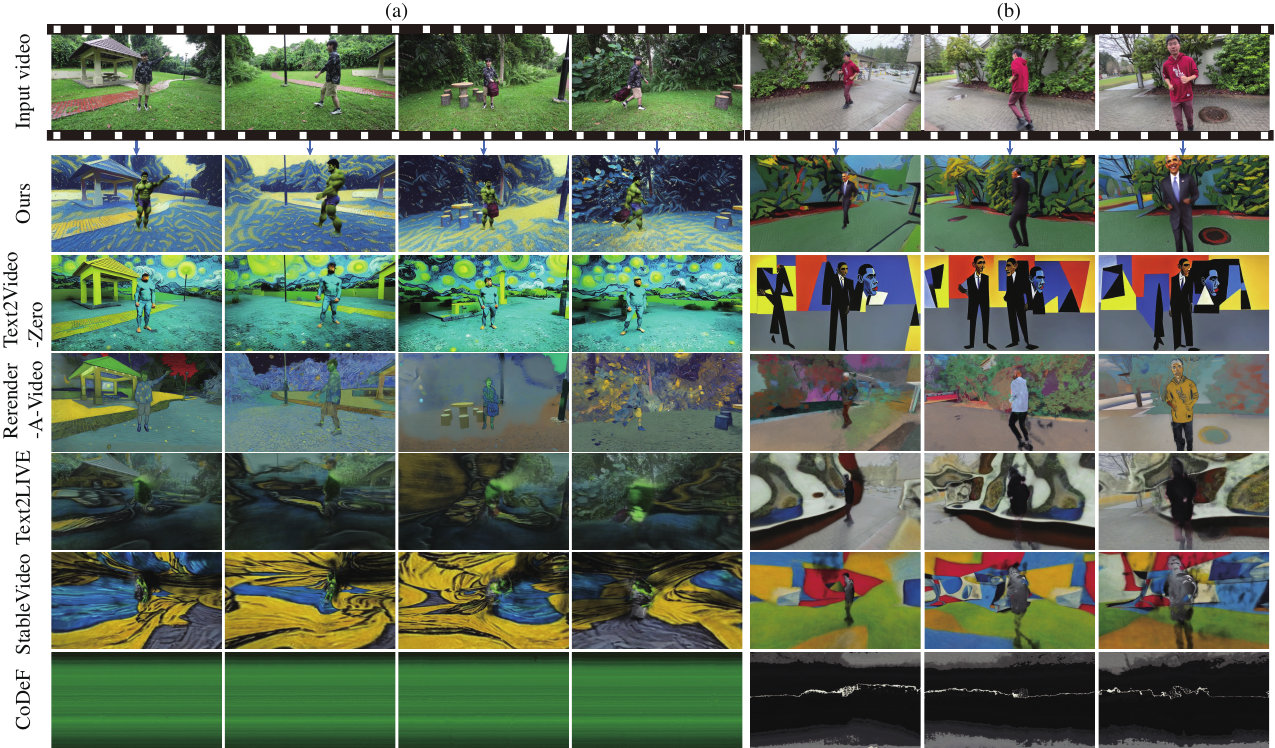}
\par\end{centering}
\vspace{-1mm}
\caption{\label{fig:results} Qualitative comparisons of \method{} against SOTA approaches on the Backpack scene (a) and Jogging scene (b).}
\vspace{-4mm}
\end{figure*}

\noindent\textbf{Qualitative Results.} We present a visual comparison of our approach against all baselines in Fig.~\ref{fig:results} for a long video (a) and a short video (b). Since both videos contain large motions and viewpoint changes, all baselines fail to edit the foreground or background and their results cannot preserve consistent structures. In contrast, our \method{} produces high-quality edited videos that accurately edit both the foreground subject and background style and maintains high temporal consistency, which largely outperforms SOTA approaches. We provide more visual comparisons of all methods, the editing time comparison of all methods, and video comparisons of all methods on $24$ editing prompts in the supplementary material.

It is worth noting that for challenging videos with large-scale motions and viewpoint changes, CoDeF~\cite{ouyang2023codef}, Text2LIVE~\cite{bar2022text2live}, and StableVideo~\cite{chai2023stablevideo} largely overfit to input video frames and learn meaningless canonical images or neural atlas, and thus cannot generate meaningful editing results. We show examples of their learned canonical images and neural atlas in the supplementary material.

\newcommand{\tablefirst}[0]{\cellcolor{myred}}

\begin{table*}[h]

\centering

\resizebox{0.75 \linewidth}{!}{
\centering
\setlength{\tabcolsep}{1.8pt}

\begin{tabular}{l||c||ccc}

\toprule
& \multicolumn{ 1 }{c||}{
  \makecell{
  \textsc{\small Metrics }
  }
}
& \multicolumn{ 3 }{c}{
  \makecell{
  \textsc{\small Human Preference}
  }
}

\\

& \multicolumn{1}{c||}{ \footnotesize CLIPScore ($\uparrow$) }
& \multicolumn{1}{c}{ \footnotesize Textual Faithfulness ($\uparrow$) }
& \multicolumn{1}{c}{ \footnotesize Temporal Consistency ($\uparrow$) }
& \multicolumn{1}{c}{ \footnotesize Overall Quality ($\uparrow$) }

\\
\hline

  Text2Video-Zero~\citep{khachatryan2023text2video}
  &$26.70$
  
  &$9.17$ v.s. $\mathbf{90.83}$ \textbf{(Ours)}
  &$21.25$ v.s. $\mathbf{78.75}$ \textbf{(Ours)}
  &$12.08$ v.s. $\mathbf{87.92}$ \textbf{(Ours)}
  
  \\ 
  Rerender-A-Video~\citep{yang2023rerender}
  &$26.11$
  
  &$6.67$ v.s. $\mathbf{93.33}$ \textbf{(Ours)}
  &$25.00$ v.s. $\mathbf{75.00}$ \textbf{(Ours)}
  &$9.58$ v.s. $\mathbf{90.42}$ \textbf{(Ours)}
  
  \\ 
  Text2LIVE~\citep{bar2022text2live}
  &$22.77$
  
  &$3.81$ v.s. $\mathbf{96.19}$ \textbf{(Ours)}
  &$26.67$ v.s. $\mathbf{73.33}$ \textbf{(Ours)}
  &$9.05$ v.s. $\mathbf{90.95}$ \textbf{(Ours)}

  \\ 
  StableVideo~\citep{chai2023stablevideo}
  &$22.02$
  
  &$4.29$ v.s. $\mathbf{95.71}$ \textbf{(Ours)}
  &$24.29$ v.s. $\mathbf{75.71}$ \textbf{(Ours)}
  &$6.19$ v.s. $\mathbf{93.81}$ \textbf{(Ours)}

  \\ 
  CoDeF~\citep{ouyang2023codef}
  &$16.77$
  
  &$1.25$ v.s. $\mathbf{98.75}$ \textbf{(Ours)}
  &$3.75$ v.s. $\mathbf{96.25}$ \textbf{(Ours)}
  &$1.25$ v.s. $\mathbf{98.75}$ \textbf{(Ours)}

  \\ 
  \method{} (Ours)
  &$\mathbf{31.31}$
  
  &--
  &--
  &--

  \\ \bottomrule

\end{tabular}
}

\vspace{-1mm}
\caption{
Quantitative comparisons of our \method{} against SOTA approaches on HOSNeRF dataset~\cite{liu2023hosnerf} and NeuMan dataset~\cite{jiang2022neuman}.
\label{tab:sota_comparison}
}
\vspace{-1mm}
\end{table*}

\noindent\textbf{Quantitative Results.} We quantify our method against
baselines through standard metrics and human preferences. We measure the textual faithfulness by computing the average CLIPScore~\cite{hessel2021clipscore} between all frames of output edited videos and corresponding text descriptions. As shown in Tab.~\ref{tab:sota_comparison}, our \method{} achieves the highest textual faithfulness score among all approaches.

\textit{Human Preference.} We show the pairwise comparing videos and textual descriptions to raters, and ask them to select their preference videos in terms of textual faithfulness, temporal consistency, and overall quality. We utilize Amazon MTurk~\footnote{https://requester.mturk.com/} to recruit $10$ participants for each comparison (Each comparison may recruit different raters), and compute their preferences over all comparisons on $24$ editing prompts. For each comparison, we show our result and one baseline result (shuffled order in questionnaires), together with textual descriptions to raters and ask their preferences. In total, we collected $1140$ comparisons over all pairwise results from $32$ different raters. As shown in Tab.~\ref{tab:sota_comparison}, we report the comparison ``$p_1\%~\text{v.s.}~p_2\%$'' where $p_1$ represents the percentage of a baseline is preferred and $p_2$ denotes our method is preferred. As evident in Tab.~\ref{tab:sota_comparison}, our method achieves the highest human preference in all aspects and outperforms all baselines by a large margin of $50\%\sim95\%$.

\begin{table}

\centering

\resizebox{1.0 \linewidth}{!}{
\centering
\setlength{\tabcolsep}{1.8pt}

\begin{tabular}{lll}
\toprule 
Ablation components & Backpack & Lab\tabularnewline
\midrule 
Full model & $\mathbf{0.756}$ & $\mathbf{0.647}$\tabularnewline
w/o Super-solution & $0.736$ & $0.645$\tabularnewline
w/o Super-solution, Rec & $0.728$ & $0.617$\tabularnewline
w/o Super-solution, Rec, 2D SDS & $0.679$ & $0.517$\tabularnewline
w/o Super-solution, Rec, 3D SDS & $0.711$ & $0.613$\tabularnewline
w/o Super-solution, Rec, 3D SDS, 2D LoRA & $0.698$ & $0.539$\tabularnewline
\bottomrule
\end{tabular}

}

\vspace{-1mm}
\caption{
Quantitative ablation results of our method for the Backpack and Lab scene (higher score means better performance).
\label{tab:ablation}
}
\vspace{-5mm}
\end{table}

\subsection{Ablation Study}

\begin{figure*}
\begin{centering}
\includegraphics[width=0.95 \linewidth]{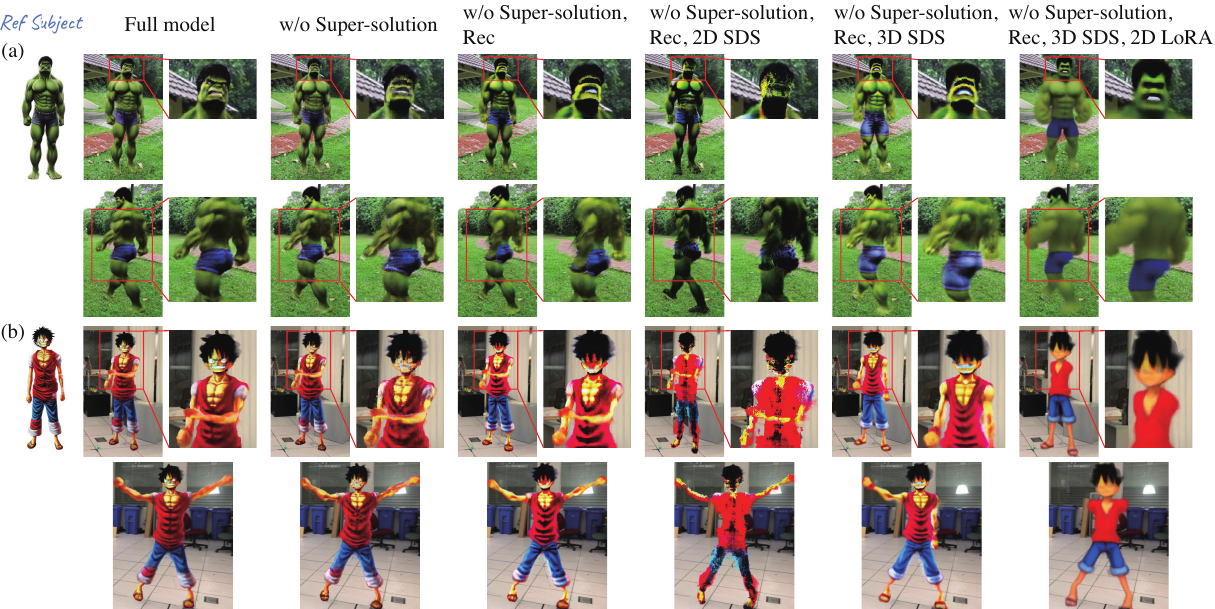}
\par\end{centering}
\vspace{-2mm}
\caption{\label{fig:ablation}Qualitative ablation results of our method on each proposed component for (a) Backpack scene and (b) Lab scene.}
\vspace{-4mm}
\end{figure*}

We conduct ablation studies on $2$ videos from HOSNeRF dataset~\cite{liu2023hosnerf} and NeuMan dataset~\cite{jiang2022neuman}. To evaluate the effectiveness of each proposed component in \method{}, we progressively ablate each component from local parts super-resolution, reconstruction loss, 2D personalized SDS, 3D SDS, and 2D personalization LoRA. To provide the quantitative results of our ablation study,  we compute the average cosine similarity between the CLIP~\cite{radford2021learning} image embeddings of all frames of output edited videos and the corresponding reference subject image. As evident in Tab.~\ref{tab:ablation}, the CLIP score progressively drops with the disabling of each component, with the full model achieving the best performances, which clearly demonstrates the effectiveness of our designs. In addition, we provide the qualitative results of our ablations in Fig.~\ref{fig:ablation}, which further demonstrates the effectiveness of our designs. More ablation results on more videos are provided in the supplementary material.

\section{Conclusion}

We introduced a novel framework of \method{} to consistently edit large-scale motion- and view-change human-centric videos. We first proposed to harness dynamic NeRF as our innovative video representation where the editing can be performed in dynamic 3D spaces and accurately propagated to the entire video via deformation fields. Then, we proposed a set of effective image-based video-NeRF editing designs, including multi-view multi-pose Score Distillation Sampling (SDS) from both the 2D personalized diffusion prior and 3D diffusion prior, reconstruction losses on the reference image, text-guided local parts super-resolution, and style transfer for 3D background spaces. Finally, extensive experiments demonstrated \method{} produced significant improvements over SOTA approaches.

\noindent\textbf{Limitations and Future Work.} Although \method{} achieves remarkable progress in video editing, its NeRF-based representation is time-consuming. Using voxel or hash grid in the video-NeRF model can largely reduce the training time and we leave it as a faithful future direction.

{
    \small
    \bibliographystyle{ieeenat_fullname}
    \bibliography{main}
}

\appendix
\newpage

\section*{Appendix}

The supplementary material is structured as follows:

\begin{itemize}
\item Sec.~\ref{sec:1} presents implementation details on the network designs and optimization parameters of \method{}.
\item Sec.~\ref{sec:3} summarizes additional comparisons and ablations of our \method{} against SOTA approaches.

\end{itemize}

Furthermore, we provide a \textbf{supplementary video} showcasing all $24$ edited video comparisons of our method against baselines, as well as 360{\textdegree} free-viewpoint renderings of edited dynamic scenes from our \method{}.

\section{Implementation Details}
\label{sec:1}

\begin{figure*}[h]
\begin{centering}
\includegraphics[width=0.8\linewidth]{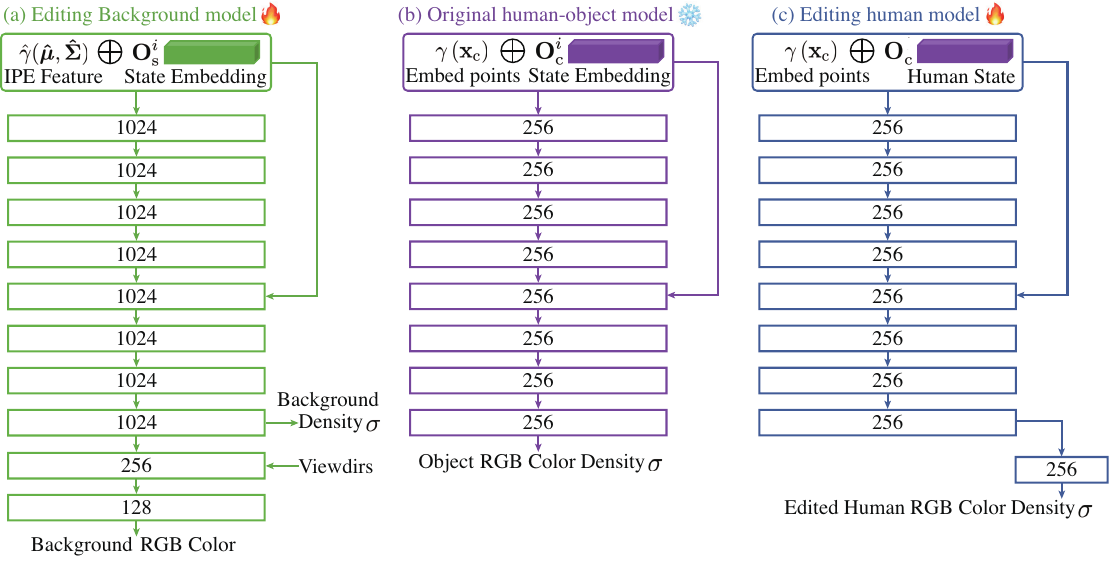}
\par\end{centering}
\vspace{-2mm}
\caption{\label{fig:networks} \method{} network designs: (a) Editing Background model, (b) Original human-object model, (c) Editing human model.}
\vspace{-2mm}
\end{figure*}

\noindent\textbf{\method{} Network Details.} As shown in Fig.~\ref{fig:networks}, we employ a 10-layer multilayer perceptron (MLP) as our state-conditional background network (a) and a 8-layer MLP as our state-conditional canonical human-object network (b). To edit the dynamic human, we establish a 9-layer canonical human network (c) where the parameters of its first 8 layers are initialized from the reconstructed human-object model (b). During optimization, we train the 3D background model (a) and the 3D dynamic human model (c) while freeze the reconstructed dynamic human-object model (b). During inference, for the source video that contains dynamic objects, we query the original dynamic human-object model (b) for the pixels within the object masks to keep the dynamic objects, while we query the edited dynamic human model (c) and edited background model for other pixels to obtain the colors and densities of edited contents. For the human-background videos, we only need to query the edited dynamic human model and edited background model to obtain the edited contents.

\noindent\textbf{Optimization Parameters.} We optimize our \method{} using Adam optimizer~\cite{kingma2014adam}. We set the learning rate for our training process as $0.0005$ with $20000$ training iterations. We balance the loss terms using the following weighting factors: $\lambda_{\mathrm{rgb}}=5,\,\lambda_{\mathrm{mask}}=0.5,\,\lambda_{\mathrm{depth}}=0.01,\,\lambda_{\mathrm{3D}}=40,\,\lambda_{\mathrm{2D}}=1.0,\,\lambda_{\mathrm{NNFM}}=1.0$. The guidance scale of the 3D diffusion prior and 2D personalized diffusion prior are set to $5$ and $20$, respectively. We conducted all our experiments on 1 NVIDIA A100 GPU, using the PyTorch~\cite{paszke2019pytorch} deep learning framework.

\begin{figure*}[h]
\begin{centering}
\includegraphics[width=0.8\linewidth]{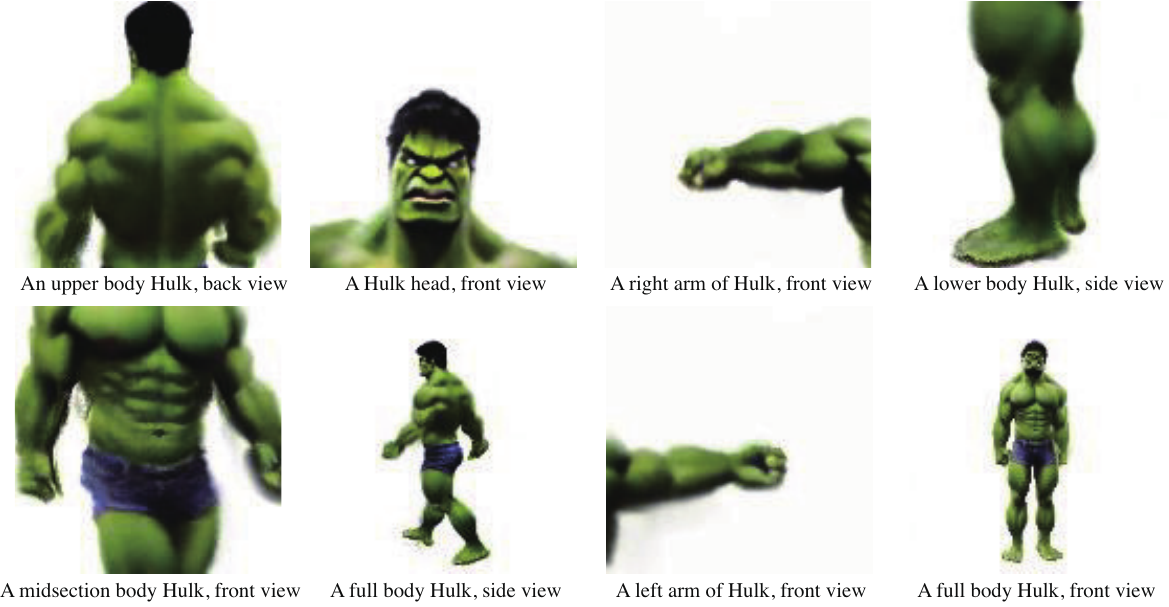}
\par\end{centering}
\vspace{-2mm}
\caption{\label{fig:superresolution} Visualization examples of text-guided local parts super-resolution sampled during training.}
\vspace{-5mm}
\end{figure*}

\noindent\textbf{Visualization of Text-guided Local Parts Super-Resolution.} 
To improve the effective resolution during training, we utilize the text-guided local parts super-resolution to render and supervise the local parts of zoom-in humans and augment with view-conditional prompts. We provide $8$ visualization examples of text-guided local parts super-resolution sampled during training in Fig.~\ref{fig:superresolution}. As shown in Fig.~\ref{fig:superresolution}, even though all figures are rendered in $\left(128\times128\right)$ resolutions, rendering local parts can largely improve the effective resolution and thus we can supervise the detailed geometry and textures of edited human body with diffusion priors.

\section{Additional Results}\label{sec:add_results}
\label{sec:3}

\begin{figure*}[h]
\begin{centering}
\includegraphics[width=0.9\linewidth]{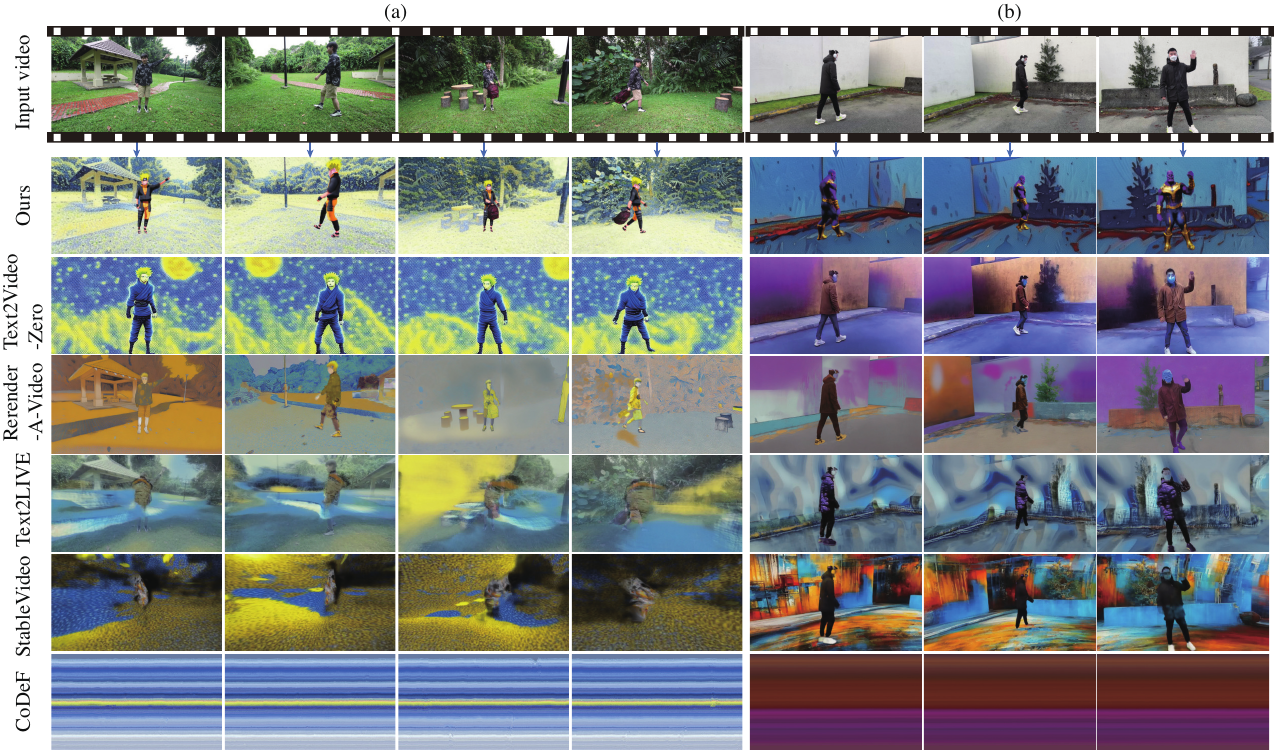}
\par\end{centering}
\vspace{-2mm}
\caption{\label{fig:results_appen1} More qualitative comparisons of \method{} against SOTA approaches on the Backpack scene (a) and Parkinglot scene (b).}
\vspace{-3mm}
\end{figure*}

\begin{figure*}[ht]
\begin{centering}
\includegraphics[width=0.9\linewidth]{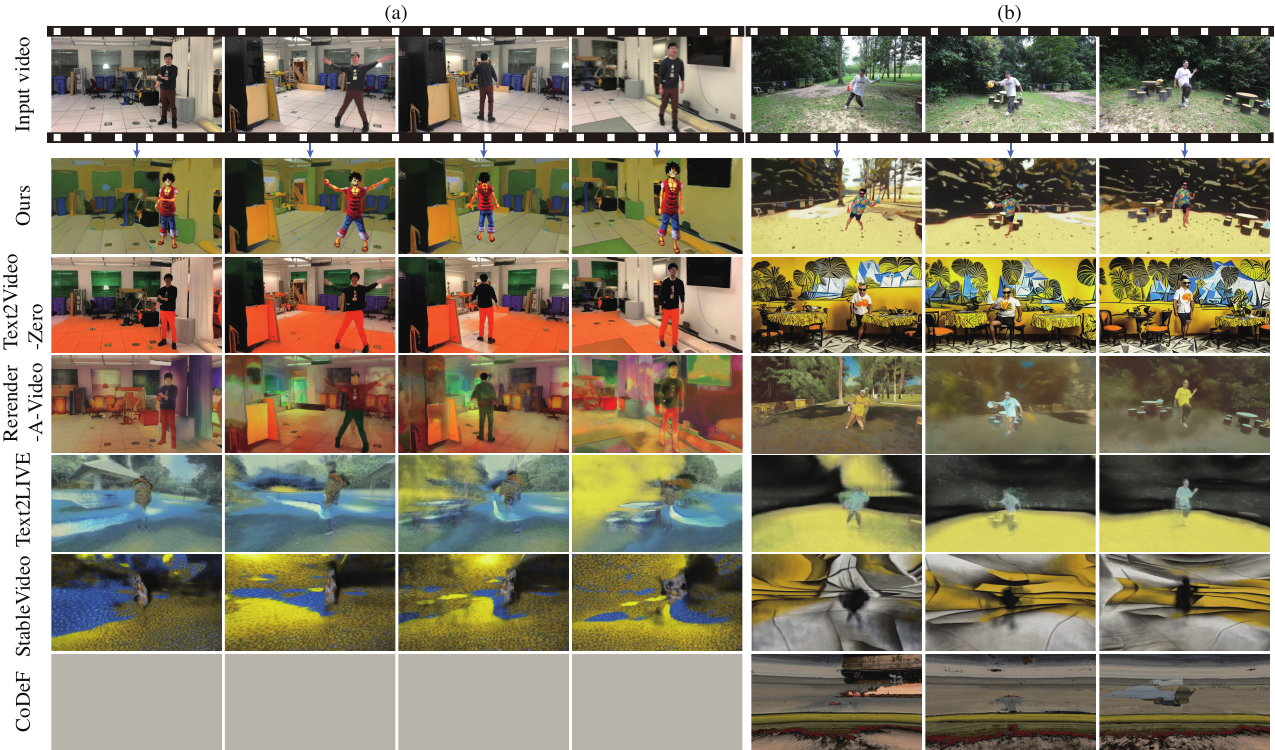}
\par\end{centering}
\vspace{-2mm}
\caption{\label{fig:results_appen2} More qualitative comparisons of \method{} against SOTA approaches on the Lab scene (a) and Dance scene (b).}
\vspace{-5mm}
\end{figure*}

\noindent\textbf{More Qualitative Results.} We present two more visual comparisons of our approach against all baselines in Fig.~\ref{fig:results_appen1} and Fig.~\ref{fig:results_appen2}. As shown in the figures, our \method{} achieves the best performances with photorealistic edited videos, which clearly demonstrates the superiority of our model against other approaches on editing large-scale motion- and view-change human-centric videos. Comparing the long (a) and short (b) video editing results of Fig.~\ref{fig:results_appen1}, we find that baseline approaches perform better on short videos than long videos, but still none of them can edit the correct subject ``Thanos'' due to the large subject motions and viewpoint changes in videos. In contrast, our \method{} produces high-quality editing results on both short and long videos. Please refer to our supplementary video for all $24$ edited video comparisons of our method against baselines.

\begin{table}
 
\centering

\resizebox{1.0 \linewidth}{!}{
\centering
\setlength{\tabcolsep}{1.8pt}

\begin{tabular}{lc}
\toprule 
Ablation components & Average CLIP Score\tabularnewline
\midrule 
Full model & $\mathbf{0.674}$ \tabularnewline
w/o Super-solution & $0.659$ \tabularnewline
w/o Super-solution, Rec & $0.650$ \tabularnewline
w/o Super-solution, Rec, 2D SDS & $0.572$ \tabularnewline
w/o Super-solution, Rec, 3D SDS & $0.641$ \tabularnewline
w/o Super-solution, Rec, 3D SDS, 2D LoRA & $0.593$ \tabularnewline
\bottomrule
\end{tabular}

}

\vspace{-2mm}
\caption{
Averaged quantitative ablation results of our method.
\label{tab:ablation_more}
}
\vspace{-7mm}
\end{table}
\noindent\textbf{Additional Ablation Results.} 
We conduct ablation studies on more videos from HOSNeRF dataset~\cite{liu2023hosnerf} and NeuMan dataset~\cite{jiang2022neuman}. To evaluate the effectiveness of each proposed component in \method{}, we progressively ablate each component from local parts super-resolution, reconstruction loss, 2D personalized SDS, 3D SDS, and 2D personalization LoRA. We observe that the model even fails to converge on some videos when we disable several components of our model. We compute the average CLIP score of all successfully edited videos in Tab.~\ref{tab:ablation_more}, where the CLIP score progressively drops with the disabling of each component, with the full model achieving the best performances, which clearly demonstrates the effectiveness of our designs.

\begin{figure*}[ht]
\begin{centering}
\includegraphics[width=1\linewidth]{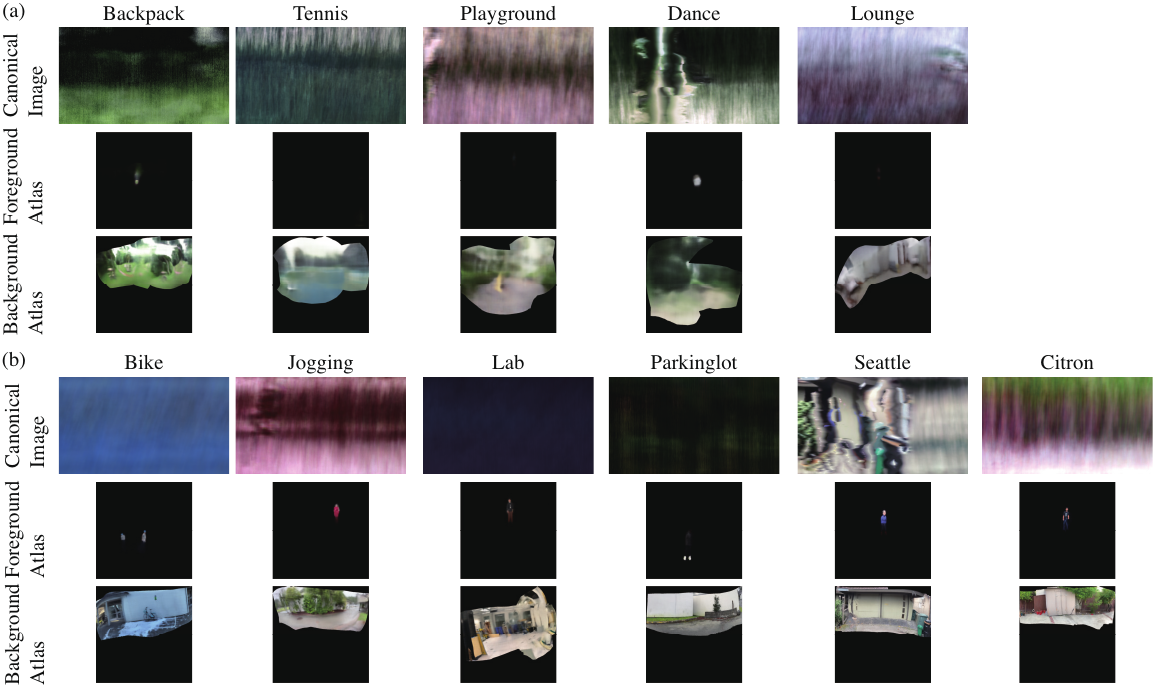}
\par\end{centering}
\vspace{-1mm}
\caption{\label{fig:canonical_appen2} Visualization of canonical images from CoDeF~\cite{ouyang2023codef}, and foreground and background atlas from Text2LIVE~\cite{bar2022text2live} and StableVideo~\cite{chai2023stablevideo} on (a) HOSNeRF dataset~\cite{liu2023hosnerf} and (b) NeuMan dataset~\cite{jiang2022neuman}.}
\vspace{-1mm}
\end{figure*}

\noindent\textbf{Visualization of Canonical Images from CoDeF~\cite{ouyang2023codef} and Atlas from Text2LIVE~\cite{bar2022text2live} and StableVideo~\cite{chai2023stablevideo}.} For challenging videos with large-scale motions and viewpoint changes, CoDeF~\cite{ouyang2023codef}, Text2LIVE~\cite{bar2022text2live}, and StableVideo~\cite{chai2023stablevideo} largely overfit to input video frames and learn meaningless canonical images or neural atlas, and thus cannot generate meaningful editing results. We show several examples of their learned canonical images~\cite{ouyang2023codef} and neural atlas~\cite{bar2022text2live,chai2023stablevideo} in Fig.~\ref{fig:canonical_appen2}, where Text2LIVE~\cite{bar2022text2live} and StableVideo~\cite{chai2023stablevideo} utilizes the same foreground and background atlas during editing. As shown in Fig.~\ref{fig:canonical_appen2}, canonical images and atlas all fail to represent the challenging large-scale motion- and view-change videos, and thus they cannot generate satisfactory editing results. In addition, the atlas performs better for short videos in NeuMan dataset~\cite{jiang2022neuman} than long videos with a better background atlas, but the foreground atlas still cannot represent the humans with large motions. In contrast, our \method{} represents videos with the dynamic NeRFs to effectively aggregate the large-scale motion- and view-change video information into a 3D dynamic human space and a 3D background space, and achieves high-quality video editing results by editing the 3D dynamic spaces.

\begin{table*}

\centering

\resizebox{0.8 \linewidth}{!}{
\centering
\setlength{\tabcolsep}{1.8pt}

\begin{tabular}{ccccccc}
\toprule 
Method & CoDeF~\cite{ouyang2023codef} & Text2Video-Zero~\cite{khachatryan2023text2video} & Rerender-A-Video~\cite{yang2023rerender} & StableVideo~\cite{chai2023stablevideo} & Text2LIVE~\cite{bar2022text2live} & \method{} (Ours)\tabularnewline
\midrule 
Time & $\sim1\,\mathrm{mins}$ & $15\,\mathrm{mins}$ & $1.2\,\mathrm{hrs}$ & $\sim1\,\mathrm{mins}$ & $\sim2\,\mathrm{hrs}$ & $7.3\,\mathrm{hrs}$\tabularnewline

\bottomrule
\end{tabular}

}

\vspace{-1mm}
\caption{
Editing operation time comparison of our method against other approaches.
\label{tab:time}
}
\vspace{-5mm}
\end{table*}

\noindent\textbf{Editing Operation Time Comparison.}
We compare the editing operation time of our \method{} against other approaches on a long video of the HOSNeRF dataset ($\left[300,\,400\right]$ frames) using a single A100 GPU in Tab.~\ref{tab:time}. Although other approaches are faster than ours, 2D-video representation-based methods such as CoDeF~\cite{ouyang2023codef}, StableVideo~\cite{chai2023stablevideo}, and Text2LIVE~\cite{bar2022text2live} cannot accurately reconstruct large-scale motion- and view-change videos and thus fail to generate meaningful editing results, as validated in Fig.~\ref{fig:canonical_appen2}. Text2Video-Zero~\cite{khachatryan2023text2video} and Rerender-A-Video~\cite{yang2023rerender} fail to edit the challenging human-centric videos with large-scale motion and viewpoint changes and their editing results are highly inconsistent. Therefore, previous approaches cannot handle the challenging human-centric videos no matter how many computation resources are provided. In contrast, our method is the first work to achieve highly consistent long-term video editing that outperforms previous approaches by a large margin of $50\%\sim95\%$ in terms of human preference, and we leave accelerating our model with voxel or hash grid representation as a faithful future direction.

\begin{figure*}[ht]
\begin{centering}
\includegraphics[width=1.0\linewidth]{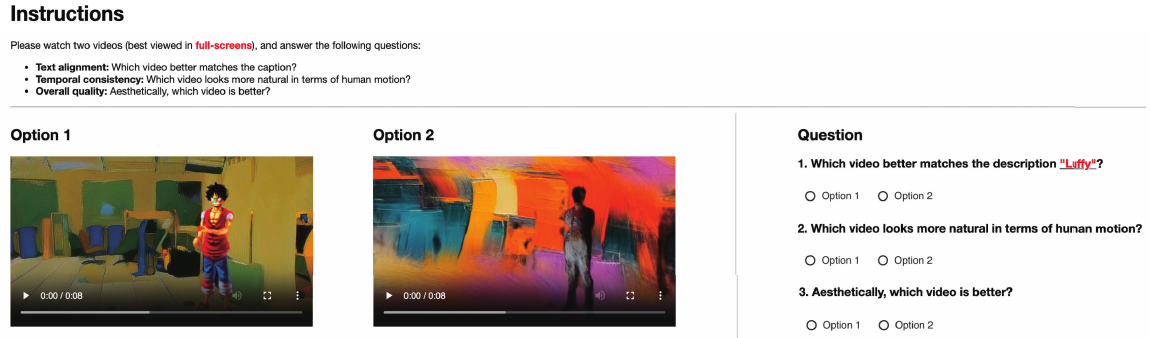}
\par\end{centering}
\vspace{-2mm}
\caption{\label{fig:question_appen} One comparison example from our questionnaires.}
\vspace{-5mm}
\end{figure*}

\noindent\textbf{Example of Human Preference Questionnaire.}
We utilize Amazon MTurk~\footnote{https://requester.mturk.com/} to recruit raters to rate our pairwise comparing videos. For each comparison, we show our result and one baseline result (shuffled order in questionnaires), together with textual descriptions to raters and ask their preferences. In total, we collected $1140$ comparisons over all pairwise results from $32$ different raters. Fig.~\ref{fig:question_appen} illustrate one comparison example in our questionnaires.

\end{document}